\documentclass[runningheads]{llncs}
\usepackage[T1]{fontenc}
\usepackage{amsmath,amssymb,amsfonts}
\usepackage{graphicx,verbatim}
\usepackage{booktabs}
\usepackage{hyperref}
\usepackage{relsize}
\usepackage{xcolor}
\usepackage{multirow}
\usepackage{marvosym}
\begin{document}

\bibliographystyle{unsrt}
\title{CIGTSurv: Clinical Information Guided Tri-modal Survival Prediction with Local Prototype Association and Global Feature Alignment}
%

\author{
Jing Dai\inst{1,2} \and
Qibin Zhang\inst{2} \and
Weiwei Zhou\inst{2} \and
Mingde Xu\inst{2} \and
Jingsong Liu\inst{3,4} \and \\
Jingdong Zhang\inst{1,2} \and
Hongming Xu\inst{1,2,5}$^\text{\Letter}$
}
\authorrunning{J. Dai et al.}
\institute{Cancer Hospital of Dalian University of Technology, Shenyang, China \and
School of Biomedical Engineering, Faculty of Medicine, \\
Dalian University of Technology, Dalian, China \and Institute of Pathology, Technical University of Munich, \\ TUM School of Medicine and Health, Munich, Germany \and Munich Center for Machine Learning (MCML), Munich, Germany \and
Key Laboratory of Integrated Circuit and Biomedical Electronic System, \\
Dalian University of Technology, Dalian, China \\
\email{\{mxu\}@dlut.edu.cn}
}
\maketitle              
\begin{abstract}
Multimodal learning has significantly advanced survival prediction by integrating pathology images with genomic data. However, clinical information, despite its critical role in reflecting a patient’s overall health, remains underutilized due to its discrete, sparse, and low-dimensional nature. Furthermore, the inherent heterogeneity across these modalities pose significant challenges in modeling cross-modal interactions. In this paper, we propose CIGTSurv, a Clinical Information Guided Tri-modal framework for Survival prediction. Specifically, we first design a holistic text template and use pretrained foundation models to transform clinical tabular data into high-dimensional tokenized embeddings. Using clinical information as an anchor, we then introduce a dual-level interaction mechanism: 1) a local prototype association (LPA) module based on cross-attention to explicitly learn token-level correspondences between different modalities, and 2) a global feature alignment (GFA) loss based on Maximum Mean Discrepancy (MMD) to implicitly enhance cross-modal distribution consistency. Extensive experiments on five TCGA cancer cohorts demonstrate that CIGTSurv achieves state-of-the-art (SOTA) survival prediction performance. Our source code is publicly available at https://github.com/Daijing-ai/CIGT-Surv.git.

\end{abstract}

\section{Introduction}
Survival prediction is a cornerstone of clinical oncology, dedicated to estimating the time-to-event for critical outcomes. Accurate prediction provides essential insights into disease progression, treatment response, and patient prognosis, ultimately guiding personalized therapeutic decision making~\cite{zhou2024multimodal}. 

With the rapid advancement of digital pathology, whole-slide images (WSIs) have been increasingly used for survival prediction. WSIs provide a comprehensive view of morphological changes at the cellular and tissue levels, offering a critical basis for survival assessment. Given the large number of patches extracted from WSIs, many methods~\cite{ilse2018attention,lu2021data,shao2021transmil,tang2024feature} adopt Multiple Instance Learning (MIL)~\cite{amores2013multiple} for efficient processing and analysis. However, pathology images alone cannot comprehensively reflect the entire process of cancer occurrence and development. Therefore, recent advances~\cite{chen2022pan,chen2021multimodal,xu2023multimodal,jaume2024modeling,song2024multimodal,zhou2024cohort} that integrate genomic data with WSIs have demonstrated improved performance, as these complementary modalities provide a more comprehensive understanding of the disease.

Nevertheless, clinical information, which provides complete context for survival prediction by enhancing the model’s understanding of a patient’s overall health~\cite{yang2025mmsurv}, have not been fully explored. This is mainly due to the discrete and low-dimensional nature of clinical tabular data, which limits their effectiveness in deep learning models that typically excel with high-dimensional features~\cite{steyaert2023multimodal}. Traditional encoding techniques, such as one-hot encoding~\cite{sedighi2021two}, struggle with discrete clinical features like T, N, and M stages, treating different categorical variables as entirely independent entities and overlooking potential intrinsic relationships between them. Although the recent SurvPGC framework~\cite{hou2025multimodal}  demonstrated the importance of incorporating clinical information to improve survival prediction via text templates and foundation models, its variable-wise prompting strategy generates isolated embedding vectors for each indicator, which overlooks the intrinsic semantic dependencies among clinical indicators. Furthermore, the inherent heterogeneity across modalities (e.g., WSIs, genomics and texts) creates a significant gap that hinders effective cross-modal integration. The overwhelming amount of WSI data can easily obscure information from other modalities, ultimately leading to information loss and suboptimal predictive performance.

To address the above challenges, we propose CIGTSurv, a Clinical Information Guided Tri-timodal framework for Survival prediction. Our main contributions are: (1) We design a holistic text template that synthesizes discrete variables into a coherent clinical description and tokenize clinical text into high-dimensional embeddings to support survival prediction. (2) We introduce a local prototype association (LPA) based on cross-attention for explicit learning of token-level correspondences, complemented by a global feature alignment (GFA) based on Maximum Mean Discrepancy (MMD) for implicit distribution alignment. The dual-level interaction mechanism ensures model to exploit both local and global relationships between modalities, thereby enabling comprehensive cross-modal interaction. (3) Extensive experiments across five cancer data cohorts demonstrate the superiority and efficiency of the proposed method compared to SOTA baselines in survival prediction.

\begin{figure}[tp]
    \centering
    \includegraphics[width=\columnwidth]{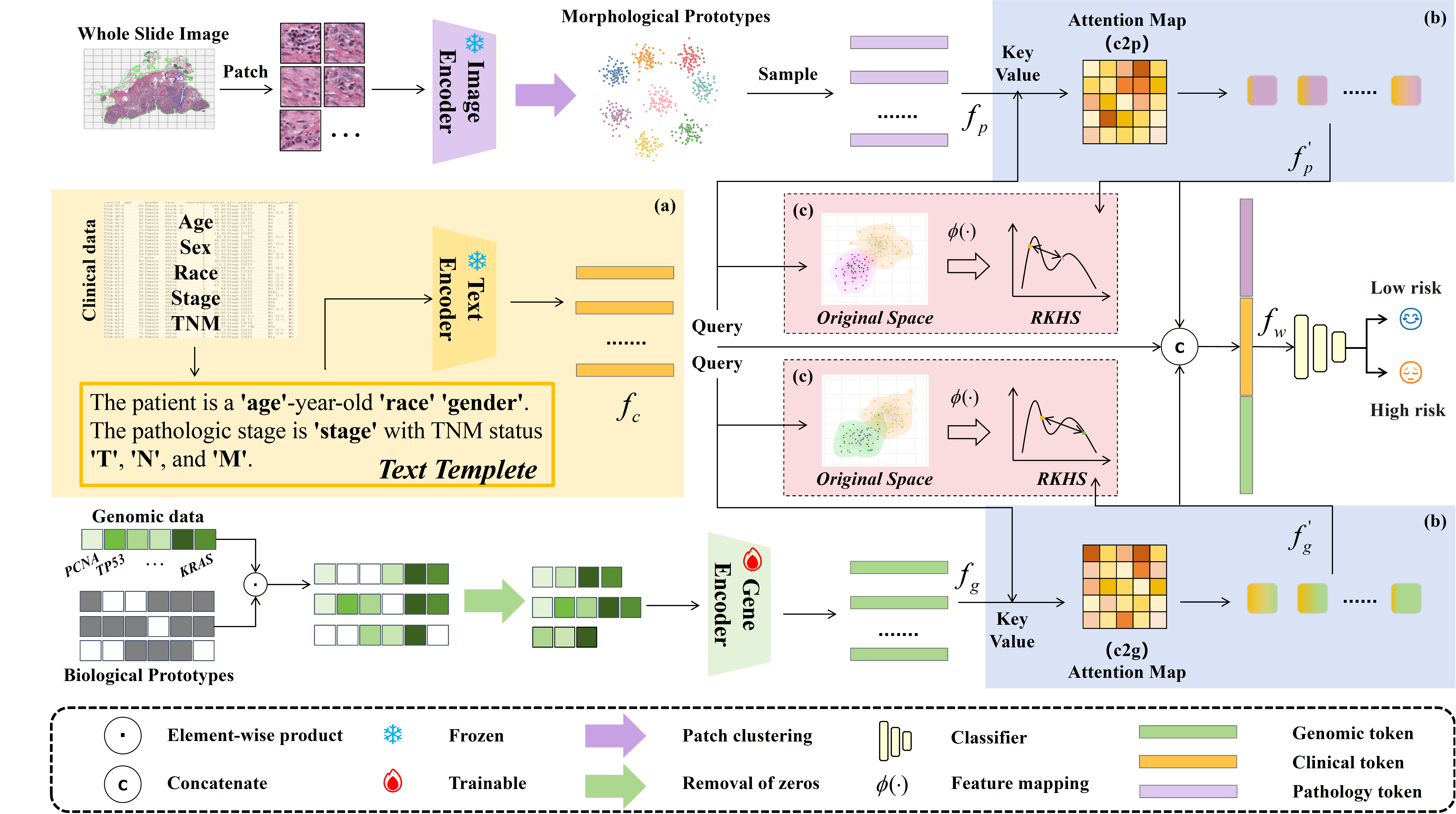}
    \caption{Overview of our proposed CIGTSurv framework featured by (a) text encoding from clinical information, (b) local prototype association and (c) global feature alignment.}
    \label{fig:overview}
\end{figure}

\section{Methods}
\subsection{Overview}
The overall architecture of the proposed CIGTSurv is shown in Fig.~\ref{fig:overview}. The framework first encodes raw signals from each modality through modality-specific encoders to generate corresponding unimodal embedding sequences $f_p$,
$f_g$, and $f_c$. It then employs a dual-level interaction mechanism consisting of a LPA module for capturing token-level correspondences and a GFA loss for enforcing distribution-level consistency. Using the clinical modality as the anchor, this mechanism yields aligned embedding sequences $f_{p}^{\prime}$ and $f_{g}^{\prime}$. These aligned unimodal embeddings, enriched with cross-modal correspondence information, are subsequently concatenated with clinical embedding $f_{c}$ to from a fused feature representation $f_{w}$. Finally, the fused tri-modal features are utilized for survival prediction.

\subsection{Text Encoding from Clinical Information}
Clinical data are inherently low-dimensional and tabular, making unified feature encoding challenging. To address this, we first design a holistic description to transform clinical information into descriptive sentences that capture contextual semantics. To ensure completeness and prognostic relevance, key patient characteristics including age, race, gender, tumor stage, and tumor TNM status are selected. For every patient, we construct a text template "\texttt{The patient is a {\textbf{\{age\}}}-year-old {\textbf{\{race\}}} {\textbf{\{gender\}}}. The pathologic stage is {\textbf{\{stage\}}} with TNM status {\textbf{\{T\}}}, {\textbf{\{N\}}}, and {\textbf{\{M\}}}}". The resulting text is encoded using BiomedBERT~\cite{gu2021domain}, resulting in a clinical embedding $I_c\in\mathbb{R}^{n_c\times d_c}$, where $n_c$ denotes the number of clinical tokens. Finally, a Multi-Layer Perceptron (MLP) projects this representation into a clinical embedding $f_c\in\mathbb{R}^{n_c\times d}$, with $d=256$.

\subsection{Prototype-based Encoding}

\noindent \textbf{Morphological Prototypes from Pathology Images.} Given a WSI, we extract $N_p$ patches of size $256 \times 256$ at $20\times$ magnification using the CLAM toolbox~\cite{lu2021data}. Each patch is encoded into a 1024-dimensional feature vector using a pre-trained pathology foundation model UNI~\cite{chen2024towards}, yielding a patch-level feature set $I_p\in\mathbb{R}^{N_p\times d_p}$. To reduce feature redundancy and computational overhead while preserving representative tissue information, we adopt a clustering-based patch selection strategy. Following prior work~\cite{yan2025pathway}, all patch embeddings are grouped into $K=50$ clusters using K-means. From each cluster, $S=10$ patches are randomly sampled, resulting in $n_p = 500$ patches per WSI. The selected patch features are then projected via a MLP to obtain the final pathology embedding $f_p \in \mathbb{R}^{n_p \times d}$.

\noindent \textbf{Biological Prototypes from Genomic Data.} The genomic data for each sample can be represented by a gene-expression vector $I_g \in \mathbb{R}^{N_G}$, where $N_G$ is the total number of genes. We aim to group these genes into $n_g=6$ pre-defined biological function sets~\cite{chen2021multimodal}, each acting as a prototype. For each set $i$ we define a binary mask $B_i \in \{0,1\}^{N_G}$, where each element indicates whether a gene is included in a given set ($1$) or not ($0$). To construct the biological prototypes, we perform an element-wise product of the gene expression vector $I_g$ with the binary set vectors ($B$)~\cite{song2024multimodal}. This operation generates a prototype representation:
\begin{equation}
    Z_{g,i} = I_g \odot B_i, \quad \forall i \in [1, n_g]
\end{equation}
The resulting vector $Z_{g,i}$ is then reduced to remove any zero entries, yielding a dense but variable-length representation (due to different gene counts across function sets). To ensure a fixed size embedding, each set is transformed into a feature by two cascaded Self-normalizing Neural Network (SNN)~\cite{klambauer2017self} layers. These grouped sets are stacked to obtain genomic embedding $f_g\in\mathbb{R}^{n_g\times d}$.

\subsection{Local Prototype Association} Using clinical embedding as anchors, We employ cross-attention to model fine-grained, token-level association across modalities. Specifically, we replace the original sequences used in traditional attention mechanisms~\cite{chen2021multimodal} with prototypes to compute interactive attention, improving efficiency and reducing computational complexity. For clinic-to-pathology association (c2p), clinical embeddings guide the aggregation of pathology features into clinic-aware pathological concepts $f_p'\in\mathbb{R}^{n_c\times d}$. Formally, let $f_c \in \mathbb{R}^{n_c \times d}$ and $f_p \in \mathbb{R}^{n_p \times d}$ denote the clinical and pathology features, respectively. The cross-attention mechanism is formulated as:
\begin{equation}
    f_{p}^{\prime} = \text{CrossAttn} \left( f_{c}, f_{p} \right) = \text{Softmax} \left( \frac{Q K^{\top}}{\sqrt{d_{k}}} \right) V,
\end{equation}
where the Query ($Q$), Key ($K$), and Value ($V$) matrices are linearly projected from the corresponding modalities:
\begin{equation}
    Q = f_{c} \mathbf{W}_{q}, \quad K = f_{p} \mathbf{W}_{k}, \quad V = f_{p} \mathbf{W}_{v},
\end{equation}
where $\mathbf{W}_{q}, \mathbf{W}_{k}, \mathbf{W}_{v} \in \mathbb{R}^{d \times d_{k}}$ are learnable projection matrices, and $d_k$ denotes the scaling dimension. A similar cross-attention operation is applied for clinic-to-genomic association (c2g), producing fused genomic features $f_g'\in\mathbb{R}^{n_c\times d}$.

\subsection{Global Feature Alignment} To implicitly align the global distributions of locally fused features with anchor features, we employ Maximum Mean Discrepancy (MMD)~\cite{gretton2012kernel}. MMD quantifies the distributional divergence between different modalities by comparing their statistics in a high-dimensional Reproducing Kernel Hilbert Space (RKHS). Specifically, given two sets of feature tockens $X = \{x_i\}_{i=1}^{N}$ and $Y = \{y_j\}_{j=1}^{N}$, where $N$ denotes the sequence length, the squared MMD distance is defined as:
\begin{equation}
M M D^2(X, Y)=\left\|\frac{1}{N} \sum_{i=1}^N \phi\left(x_i\right)-\frac{1}{N} \sum_{j=1}^N \phi\left(y_j\right)\right\|_{\mathcal{H}}^2,
\end{equation}
where $\phi(\cdot)$ represents the feature mapping to the RKHS $\mathcal{H}$. In practice, this distance is efficiently computed via the kernel trick:
\begin{equation}
M M D^2(X, Y)=\frac{1}{N^2} \sum_{i=1}^N \sum_{i^{\prime}=1}^N k\left(x_i, x_{i^{\prime}}\right)+\frac{1}{N^2} \sum_{j=1}^N \sum_{j^{\prime}=1}^N k\left(y_j, y_{j^{\prime}}\right)-\frac{2}{N^2} \sum_{i=1}^N \sum_{j=1}^N k\left(x_i, y_j\right),
\end{equation}
where we employ a Gaussian kernel $k(x, y)=\exp \left(-\frac{\|x-y\|_2^2}{2 \sigma^2}\right)$ with bandwidth $\sigma$.

For the fused pathology features $f_p^{\prime}$, genomic features $f_g^{\prime}$, and clinical anchor features $f_c$, we define the global alignment loss as the cumulative of MMD distances across modality pairs:
\begin{equation}
L_{\text {align }}=MMD^2\left(f_p^{\prime}, f_c\right)+MMD^2\left(f_g^{\prime}, f_c\right).
\end{equation}
Minimizing this loss forces the model to bridge the distribution gap between the sets of local tokens across modalities. By integrating explicit local token-level associations with implicit global feature alignment, our framework facilitates multi-granular fusion and effective cross-modal interaction.

\subsection{Feature Aggregation and Survival Prediction}
For a patient, we apply mean pooling to the clinical features $f_c$, aligned pathology features $f_p^{\prime}$ and aligned genomic features $f_g^{\prime}$ and concatenate these three feature sequences to get the final tri-modal feature vector $f_{w}$. This vector is subsequently passed into a linear classifier to generate the final survival prediction. Following prior studies~\cite{chen2021multimodal,jaume2024modeling}, we employ the Negative Log-Likelihood (NLL) loss function, as  \( L_{\text {surv }}\). The total loss for end-to-end model training is then calculated with a weight factor $\lambda$:
\begin{equation}
L=L_{\text {surv }}+\lambda L_{\text {align }}.
\end{equation}

\section{Experiments}
\subsection{Datasets and Implementation Details}
We conducted experiments on five cancer cohorts derived from The Cancer Genome Atlas (TCGA)\footnote{https://portal.gdc.cancer.gov/}, including Bladder Urothelial Carcinoma (BLCA, n=337), Breast Invasive Carcinoma (BRCA, n=968), Colon and Rectum Adenocarcinoma (COADREAD, $n$ = 252), Liver Cancer (LIHC, $n$ = 311) and Kidney Papillary Cell Carcinoma (KIRP, $n$ = 261). During preprocessing, incomplete cases were excluded; therefore, all included cases contain complete age, race, gender, stage, and TNM information. We employed 5-fold cross-validation to evaluate our model and other compared methods. The mean Concordance index (C-index) with its standard deviation (std) are reported for a comprehensive performance comparison.

During each fold of cross-validation, models were trained for up to 100 epochs with early stopping based on validation C-index. A batch size of one was used, and gradients were accumulated over 32 iterations prior to backpropagation. Optimization was performed using the Adam optimizer with a learning rate of $2\times10^{-4}$ and a weight decay of $1\times10^{-5}$. The Gaussian kernel bandwidth $\sigma$ is computed using the median heuristic. The alignment weight $\lambda$ is selected from {0.01, 0.05, 0.1, 0.5, 1}, with dataset-specific values: BLCA 0.01, BRCA 1, COADREAD 0.1, LIHC 0.01, and KIRP 0.5. To address class imbalance across tasks, a weighted sampling strategy was applied during training. All experiments were implemented in PyTorch and executed on a single NVIDIA RTX 4090 GPU.

\begin{table*}[!tb]
\caption{C-index ($\uparrow$) performance (mean ± std) over five datasets. The \textbf{best} and the \underline{second best} results are highlighted.}
\label{table:comparison-merged}
\centering
\resizebox{\textwidth}{!}{
\begin{tabular}{ccl|ccccc|c}
    \toprule
 \multicolumn{3}{c|}{\textbf{Model}} &\textbf{BLCA} & \textbf{BRCA} & \textbf{COADREAD} & \textbf{LIHC} & \textbf{KIRP} & \textbf{Mean} \\
    \midrule
        && Cox$_{(age, sex, stage)}$~\cite{breslow1975analysis} & 0.645 \textsmaller{$\pm$ 0.054} & \textbf{0.772 \textsmaller{$\pm$ 0.046}} & 0.724 \textsmaller{$\pm$ 0.138} & 0.627 \textsmaller{$\pm$ 0.044} & 0.832 \textsmaller{$\pm$ 0.089} & 0.720 \\ 
    \midrule

    \multicolumn{2}{c}{\multirow{3}{*}{\rotatebox[origin=c]{90}{\textbf{Geno.}}}}&MLP  &   0.657\textsmaller{$\pm$ 0.008} & 0.647 \textsmaller{$\pm$ 0.060} & 0.604 \textsmaller{$\pm$ 0.098} & 0.621 \textsmaller{$\pm$ 0.023} & 0.823 \textsmaller{$\pm$ 0.135} & 0.670 \\ 
    &&SNN~\cite{klambauer2017self}  &   0.686 \textsmaller{$\pm$ 0.036} & 0.666 \textsmaller{$\pm$ 0.027} & 0.608 \textsmaller{$\pm$ 0.072} & 0.630 \textsmaller{$\pm$ 0.032} & 0.866 \textsmaller{$\pm$ 0.102} & 0.691 \\ 
    &&SNNTrans~\cite{klambauer2017self}  &   0.683 \textsmaller{$\pm$ 0.018} & 0.674 \textsmaller{$\pm$ 0.048} & 0.716 \textsmaller{$\pm$ 0.121} & 0.632 \textsmaller{$\pm$ 0.058} & 0.862 \textsmaller{$\pm$ 0.109} & 0.713 \\ 
    \midrule

    \multicolumn{2}{c}{\multirow{4}{*}{\rotatebox[origin=c]{90}{\textbf{Patho.}}}}& ABMIL~\cite{ilse2018attention}  & 0.688 \textsmaller{$\pm$ 0.047} & 0.711 \textsmaller{$\pm$ 0.046} & 0.764 \textsmaller{$\pm$ 0.063} & 0.733 \textsmaller{$\pm$ 0.044} & 0.873 \textsmaller{$\pm$ 0.063} & 0.754 \\ 
    &&CLAM-MB~\cite{lu2021data} & 0.664 \textsmaller{$\pm$ 0.039} & 0.697 \textsmaller{$\pm$ 0.040} & 0.804 \textsmaller{$\pm$ 0.088} & 0.728 \textsmaller{$\pm$ 0.024} & 0.847 \textsmaller{$\pm$ 0.067} & 0.753 \\ 
    &&TransMIL~\cite{shao2021transmil}   & 0.699 \textsmaller{$\pm$ 0.049} & 0.674 \textsmaller{$\pm$ 0.024} & 0.727 \textsmaller{$\pm$ 0.107} & 0.721 \textsmaller{$\pm$ 0.036} & 0.786 \textsmaller{$\pm$ 0.136} & 0.721 \\ 
    &&RRTMIL~\cite{tang2024feature}   & 0.639 \textsmaller{$\pm$ 0.030} & 0.702 \textsmaller{$\pm$ 0.013} & 0.783 \textsmaller{$\pm$ 0.088} & 0.722 \textsmaller{$\pm$ 0.038} & 0.870 \textsmaller{$\pm$ 0.068} & 0.743 \\ 
    \midrule
    
    \multicolumn{2}{c}{\multirow{6}{*}{\rotatebox[origin=c]{90}{\textbf{Dual.}}}}&Porpoise~\cite{chen2022pan}  & 0.690 \textsmaller{$\pm$ 0.041} & 0.695 \textsmaller{$\pm$ 0.056} & 0.765 \textsmaller{$\pm$ 0.106} & 0.740 \textsmaller{$\pm$ 0.046} & 0.872 \textsmaller{$\pm$ 0.090} & 0.752 \\ 
    &&MCAT~\cite{chen2021multimodal} & 0.670 \textsmaller{$\pm$ 0.020} & 0.706 \textsmaller{$\pm$ 0.057} & 0.788 \textsmaller{$\pm$ 0.072} & 0.735 \textsmaller{$\pm$ 0.047} & {0.905 \textsmaller{$\pm$ 0.038}} & 0.761 \\ 
    &&MOTCat~\cite{xu2023multimodal} & 0.702 \textsmaller{$\pm$ 0.016} & 0.720 \textsmaller{$\pm$ 0.054} & \underline{0.815 \textsmaller{$\pm$ 0.068}} & 0.736 \textsmaller{$\pm$ 0.046} & 0.901 \textsmaller{$\pm$ 0.060} & \underline{0.775} \\ 
    &&SurvPath~\cite{jaume2024modeling} &  \underline{0.712 \textsmaller{$\pm$ 0.037}} & 0.707 \textsmaller{$\pm$ 0.028} & 0.774 \textsmaller{$\pm$ 0.087} & 0.730 \textsmaller{$\pm$ 0.049} & 0.891 \textsmaller{$\pm$ 0.065} & 0.763 \\ 
    &&$MMP_{Trans.}$~\cite{song2024multimodal}  & 0.704 \textsmaller{$\pm$ 0.036} & 0.701 \textsmaller{$\pm$ 0.024} & 0.738 \textsmaller{$\pm$ 0.124} & 0.688 \textsmaller{$\pm$ 0.044} & 0.887 \textsmaller{$\pm$ 0.086} & 0.744 \\ 
    &&CCL~\cite{zhou2024cohort} & 0.691 \textsmaller{$\pm$ 0.022} & 0.660 \textsmaller{$\pm$ 0.036} &0.763 \textsmaller{$\pm$ 0.063} & 0.630 \textsmaller{$\pm$ 0.062} & 0.814 \textsmaller{$\pm$ 0.155} & 0.712 \\ 
    \midrule
    \multicolumn{2}{c}{\multirow{4}{*}{\rotatebox[origin=c]{90}{\textbf{Tri.}}}}&Concat  & 0.701 \textsmaller{$\pm$ 0.014} & 0.683 \textsmaller{$\pm$ 0.021} & 0.782 \textsmaller{$\pm$ 0.094} & \underline{0.747 \textsmaller{$\pm$ 0.034}} & 0.873 \textsmaller{$\pm$ 0.067} & 0.757 \\ 
    &&Pathomic~\cite{chen2020pathomic} & 0.699 \textsmaller{$\pm$ 0.037} & 0.693 \textsmaller{$\pm$ 0.045} & 0.701 \textsmaller{$\pm$ 0.065} & 0.725 \textsmaller{$\pm$ 0.041} & 0.885 \textsmaller{$\pm$ 0.087} & 0.741 \\ 
    &&SurvPGC~\cite{hou2025multimodal} & 0.689 \textsmaller{$\pm$ 0.021} & 0.722 \textsmaller{$\pm$ 0.040} & 0.792 \textsmaller{$\pm$ 0.070} & 0.724 \textsmaller{$\pm$ 0.055} & 0.881 \textsmaller{$\pm$ 0.075} & 0.762 \\ 
    &&\textbf{CIGTSurv (Ours)} & \textbf{0.713 \textsmaller{$\pm$ 0.026}} & \underline{0.748 \textsmaller{$\pm$ 0.040}} & \textbf{0.816 \textsmaller{$\pm$ 0.081}} & \textbf{0.749 \textsmaller{$\pm$ 0.038}} & \textbf{0.913 \textsmaller{$\pm$ 0.045}} & \textbf{0.788} \\
    \bottomrule
\end{tabular}
}
\end{table*}

\subsection{Performance Evaluation}
\noindent \textbf{Comparisons with SOTA methods.} We compared our model in five settings: clinic-only (Cox model~\cite{breslow1975analysis}), pathology-only (ABMIL~\cite{ilse2018attention}, CLAM-MB~\cite{lu2021data}, TransMIL~\cite{shao2021transmil}, and RRTMIL~\cite{tang2024feature}), genomic-only (MLP, SNN~\cite{klambauer2017self}, and SNNTrans~\cite{klambauer2017self}), dual-modal (Porpoise~\cite{chen2022pan}, MCAT~\cite{chen2021multimodal}, MOTCat~\cite{xu2023multimodal}, SurvPath~\cite{jaume2024modeling}, $MMP_{Trans.}$~\cite{song2024multimodal} and CCL~\cite{zhou2024cohort}) and tri-modal (Concat, Pathomic~\cite{chen2020pathomic} and SurvPGC~\cite{hou2025multimodal}). Table~\ref{table:comparison-merged} presents the comparative results, where our proposed CIGTSurv achieves an average C-index of 78.8\%, surpassing all other comparative methods. The superior performance is mainly attributed to its dual-level mechanism, which effectively bridges the modality gap between clinical information, pathology images, and genomic data via the combined use of cross-attention and MMD loss. While most multimodal approaches surpass unimodal baselines, CIGTSurv further excels by utilizing clinical prompts to guide the extraction of prognostic cues from high-dimensional pathology and genomic signatures. Furthermore, compared to existing tri-modal architectures, our model significantly enhances interaction efficiency by prototype construction, achieving a superior balance between computational overhead and predictive accuracy.

\begin{figure}[tb]
    \centering
    \includegraphics[width=\columnwidth]{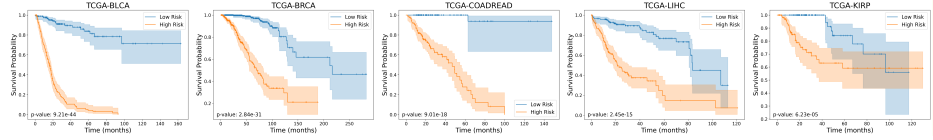}
    \caption{Kaplan-Meier (KM) curves of the proposed model on five cancer datasets, where patient stratification of low risk (\textcolor[RGB]{31,119,180}{blue}) and high risk (\textcolor[RGB]{255,127,14}{orange}) are presented.}
    \label{fig:KM}
\end{figure}

\noindent \textbf{Patient Stratification.} To validate the discriminative ability of our model, we performed Kaplan-Meier survival analysis and log-rank test by stratifying patients into high- and low-risk groups according to the median predicted risk scores, as illustrated in Fig.~\ref{fig:KM}. Statistically significant differences ($p<0.05$) were observed across all datasets, confirming the effectiveness of our model in patient risk stratification.

\begin{table*}[t]
\caption{Ablation results of our method}
\label{table:ablation}
\centering
\resizebox{\textwidth}{!}{
\begin{tabular}{l|ccccc|c}
    \toprule
    \textbf{Module} & \textbf{BLCA} & \textbf{BRCA} & \textbf{COADREAD} & \textbf{LIHC} & \textbf{KIRP} & \textbf{Mean} \\
    \midrule
    w/o Text Embed. & 0.704\textsmaller{$\pm$ 0.027} & 0.744\textsmaller{$\pm$ 0.090} & 0.779\textsmaller{$\pm$ 0.078} & 0.730\textsmaller{$\pm$ 0.038} & 0.909\textsmaller{$\pm$ 0.044} & 0.773 \\
    w/o Proto. & 0.691\textsmaller{$\pm$ 0.025} & 0.654\textsmaller{$\pm$ 0.045} & 0.706\textsmaller{$\pm$ 0.119} & 0.707\textsmaller{$\pm$ 0.072} & 0.907\textsmaller{$\pm$ 0.031} & 0.733 \\
    \midrule
    w/o LPA & 0.709\textsmaller{$\pm$ 0.032} & 0.692\textsmaller{$\pm$ 0.045} & 0.768\textsmaller{$\pm$ 0.106} & 0.737\textsmaller{$\pm$ 0.021} & 0.891\textsmaller{$\pm$ 0.060} & 0.759 \\
    w/o GFA & 0.690\textsmaller{$\pm$ 0.032} & 0.691\textsmaller{$\pm$ 0.049} & 0.801\textsmaller{$\pm$ 0.085} & 0.742\textsmaller{$\pm$ 0.030} & 0.881\textsmaller{$\pm$ 0.070} & 0.761 \\
    w/o Both & 0.694\textsmaller{$\pm$ 0.044} & 0.692\textsmaller{$\pm$ 0.016} & 0.769\textsmaller{$\pm$ 0.091} & 0.736\textsmaller{$\pm$ 0.021} & 0.860\textsmaller{$\pm$ 0.074} & 0.750 \\
    \midrule
    w/o Clinic & 0.709\textsmaller{$\pm$ 0.032} & 0.712\textsmaller{$\pm$ 0.022} & 0.766\textsmaller{$\pm$ 0.111} & 0.736\textsmaller{$\pm$ 0.033} & 0.883\textsmaller{$\pm$ 0.057} & 0.761 \\
    w/o Pathology & 0.664\textsmaller{$\pm$ 0.045} & 0.678\textsmaller{$\pm$ 0.027} & 0.706\textsmaller{$\pm$ 0.141} & 0.668\textsmaller{$\pm$ 0.056} & 0.884\textsmaller{$\pm$ 0.114} & 0.720 \\
    w/o Genomic & 0.690\textsmaller{$\pm$ 0.026} & 0.682\textsmaller{$\pm$ 0.063} & 0.795\textsmaller{$\pm$ 0.096} & 0.745\textsmaller{$\pm$ 0.027} & 0.892\textsmaller{$\pm$ 0.066} & 0.761 \\
    \midrule
    \textbf{CIGTSurv(Ours)} & \textbf{0.713\textsmaller{$\pm$ 0.026}} & \textbf{0.748\textsmaller{$\pm$ 0.040}} & \textbf{0.816\textsmaller{$\pm$ 0.081}} & \textbf{0.749\textsmaller{$\pm$ 0.038}} & \textbf{0.913\textsmaller{$\pm$ 0.045}} & \textbf{0.788} \\
    \bottomrule
\end{tabular}
}
\end{table*}

\begin{figure}[tp]
    \centering
    \includegraphics[width=\columnwidth]{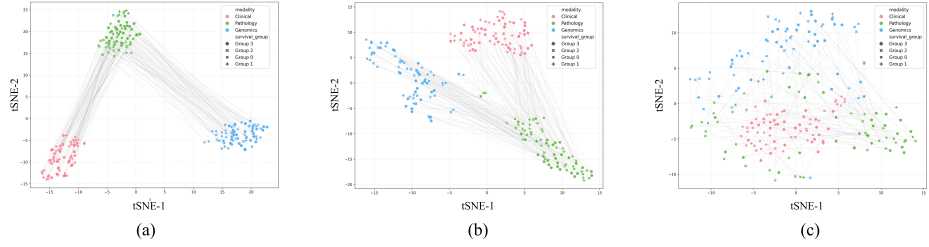}
    \caption{T-SNE visualization of the feature distributions. (a) Feature distributions in the baseline. (b) Feature distributions after incorporating LPA. (c) Feature distributions by our proposed CIGTSurv framework.}
    \label{fig:tSNE}
\end{figure}

\subsection{Ablation Studies}
We conduct comprehensive ablation studies from four aspects to evaluate our proposed method, as shown in Table~\ref{table:ablation}.

\noindent \textbf{Effectiveness of Text Encoding.} To assess the impact of different encoding strategies for clinical information, we transformed all clinical variables into one-hot encoded vectors~\cite{liang2026multi}. This approach (i.e., w/o Text Embed.) yielded an overall C-index of 0.773, which is lower than that achieved using our text-based clinical embeddings. This indicates that directly encoding clinical variables in a one-hot format may hinder the model from learning effective semantic relationships and introduce spurious associations among features.

\noindent \textbf{Effectiveness of Prototypes.} We replaced the morphological and biological prototypes with all patch tokens and genomic data (i.e., w/o Proto.). The C-index consistently dropped across all datasets, indicating that compact prototypes improve the efficiency of cross-modal information interaction.

\noindent \textbf{Effectiveness of Proposed Components.} We evaluated the contributions of local prototype association (LPA) module and global feature alignment (GFA) loss. Removing either component led to a performance drop (see Table~\ref{table:ablation}). Furthermore, we employed t-SNE~\cite{van2008visualizing} to visualize feature distributions at different stages of our framework, as shown in Fig ~\ref{fig:tSNE}. In the baseline (Fig. ~\ref{fig:tSNE}a), features form well-separated clusters, indicating a severe modality gap. With the gradual introduction of cross-attention (Fig. ~\ref{fig:tSNE}b) and MMD loss (Fig. ~\ref{fig:tSNE}c), these clusters gradually converge toward a shared latent space.

\noindent \textbf{Effectiveness of Tri-modal Fusion.} Lastly, we conducted modality ablation experiments by removing one modality in turn. Excluding the pathology modality caused a marked performance drop, reflecting the strong correlation between histopathology and survival outcomes. The highest performance was achieved when all three modalities were integrated, confirming the effectiveness of our tri-modal fusion approach.

\section{Conclusion}
In this work, we propose CIGTSurv, a novel Clinical Information Guided Tri-modal framework for cancer survival prediction. By converting clinical tabular data into holistic, semantic-rich text embeddings, our framework achieves seamless integration of clinical insights with pathology and genomic modalities. The proposed dual-level cross-modal interaction mechanism—comprising local prototype association and global distribution alignment—effectively mitigates the heterogeneous modality gap, yielding a comprehensive and highly discriminative representation. The results highlight the framework’s efficiency and potential to guide future multimodal learning research in oncology.

\subsubsection{\ackname}
This work was supported in part by the Liaoning Province Science and Technology Joint Program (No. 2024-MSLH-065), the Fundamental Research Funds for Central Universities (No. DUT25Z2514). 

\subsubsection{\discintname}
The authors have no competing interests to declare that are relevant to the content of this article.

%
%
%
%
\bibstyle{splncs04}
\bibliography{ref}

\end{document}